\documentclass{article}
\usepackage{spconf,amsmath,graphicx,hyperref}
\usepackage{amssymb}

\usepackage{booktabs,multirow,amsfonts}

\usepackage{enumitem}

\title{HMVLA: Hyperbolic Multimodal Fusion for Vision-Language-Action Models}
\name{Kun Wang$^{1,2}$ \qquad Xiao Feng$^{1,2}$  \qquad Mingcheng Qu$^{1,2}$ \qquad Tonghua Su$^{1,2,3}$ \sthanks{Corresponding author}}
  
  \address{$^{1}$ Harbin Institute of Technology, Harbin, China \\
  $^{2}$ Chongqing Research Institute of HIT, Chongqing, China\\ 
  $^{3}$ Guangdong Laboratory of Artificial Intelligence and Digital Economy (SZ), Shenzhen, China}

\begin{document}
%\ninept
%
\maketitle

\begin{abstract}
% Vision-Language-Action (VLA) models have recently shown great potential in bridging multimodal perception with robotic control. However, existing frameworks often struggle to capture hierarchical semantic relations in visual features and to generalize across diverse task demands. In this work, we propose an enhanced VLA framework that integrates hyperbolic embedding and Mixture-of-Experts (MoE) representation learning. Specifically, we introduce a hyperbolic entailment loss into the image encoder, which encourages the preservation of hierarchical and semantic structures in visual representations. Furthermore, we replace the feed-forward layers in the Q-Former module with a Mixture-of-Experts network, enabling dynamic routing and more flexible cross-modal reasoning. Experimental results on benchmark tasks demonstrate that our method significantly improves accuracy compared to conventional VLA baselines. These findings highlight the effectiveness of combining hyperbolic representation learning with MoE-enhanced architectures, providing new insights for advancing multimodal understanding and action decision-making. 
Vision-Language-Action (VLA) models have recently shown great potential in bridging multimodal perception with robotic control. However, existing methods often rely on direct fine-tuning of pre-trained Vision-Language Models (VLMs), feeding semantic and visual features directly into a policy network without fully addressing the unique semantic alignment challenges in the VLA domain. In this paper, we propose HMVLA, a novel VLA framework that exploits the inherent hierarchical structures in vision and language for comprehensive semantic alignment. Unlike traditional methods that perform alignment in Euclidean space, our HMVLA embeds multimodal features in hyperbolic space, enabling more effective modeling of the hierarchical relationships present in image–text data. Furthermore, we introduce a sparsely gated Mixture-of-Experts (MoE) mechanism tailored for semantic alignment, which enhances multimodal comprehension between images and text while improving efficiency. Extensive experiments demonstrate that HMVLA surpasses baseline methods in both accuracy and generalization. In addition, we validate its robustness by reconstructing datasets to further test cross-domain adaptability.

\end{abstract}

\begin{keywords}
Vision-Language-Action, Embodied AI, Hyperbolic space, Mixture-of-Experts
\end{keywords}
%

% \begin{figure*}[!t]\centering
% 	\includegraphics[width=\linewidth]{image/struction.pdf}
% 	\caption{High-resolution fine-grained image generation architecture diagram. The model incorporates the “Tiered Embedder" shown in the bottom right to introduce superclass information into the model, introduces the ProAttention mechanism to enhance training efficiency, and integrates the concept of super-resolution during the denoising process to generate high-quality fine-grained image pixels.}
%     \label{fig:struction}
% \end{figure*}

% \begin{figure}[t]\centering
% 	\includegraphics[width=\linewidth]{image/stru.jpg}
% 	\caption{Comparison between the ProAttention mechanism and Attention mechanism in image generation.}
%     \label{pic:pro_attention}
% \end{figure}

\section{Introduction}
\label{sec:intro}
The rapid development of Large-Language-Models (LLMs) and Vision-Language-Models (VLMs)~\cite{achiam2023gpt,dubey2024llama,sun2025llapa,zang2025sage} has significantly advanced Vision-Language-Action (VLA) models in robotics \cite{brohan2022rt,belkhale2024rt,zhou2025vision,zhen20243d,black2024pi_0,intelligence2025pi_}. By leveraging the powerful comprehension capabilities of pre-trained models, VLA systems encode visual and linguistic modalities into feature tokens, which are then fed into a robotic policy network to enable end-to-end action generalization~\cite{ma2024survey,zhang2024vision}.

% However, directly passing visual and language features into the policy network for precise robot control presents significant challenges, particularly in achieving robust generalization of actions in end-to-end frameworks\cite{firoozi2025foundation,liang2022code}.

Existing VLA models, such as RT-2\cite{zitkovich2023rt} and OpenVLA\cite{kim2024openvla}, typically fine-tune pre-trained VLMs on robot datasets, directly mapping visual and linguistic features to robot control actions. However, this strategy often disrupts the intrinsic consistency of visual and semantic features. In particular, the hierarchical structure of semantics and vision embedded in robot datasets may become perturbed. For example, when training on a ``grasp the cup'' task where the background is a white table and target is a blue mug, the model may learn spurious correlations (\textit{e.g.,} associating ``white'' with ``background'' and ``blue'' with ``cup''), instead of grounding the semantics of the action ``grasp'' and the object ``cup''.

This mismatch between execution semantics and action outcomes highlights a core limitation of current VLA approaches. The phenomenon also echoes broader findings in vision-language research: fine-tuning linguistically aligned visual encoders can lead to overfitting, ultimately failing to preserve the hierarchical semantic–visual structures necessary for robust reasoning and control~\cite{zitkovich2023rt,kim2024openvla}. 
Consequently, this weakens the mapping from visual–semantic understanding to robotic actions, leading to diminished execution capabilities. By contrast, models such as CLIP~\cite{radford2021learning} have demonstrated powerful vision-language alignment capabilities and exhibit excellent performance across various tasks. This highlights an important direction for VLA research: how to adapt pre-trained models in a way that preserves the hierarchical structure of vision and semantics~\cite{zeng2022extensive}, ensuring that fine-tuning in the robotics domain enables more faithful grounding and accurate prediction of next-step actions.

% In response to this challenge, we propose the HMVLA model. Unlike current VLA approaches, which incorporate prior knowledge from generative models in other domains or introduce techniques like chain-of-thought reasoning to facilitate model-based image inference, our model fine-tunes the Vision-Language Model (VLM) while leveraging the geometric properties of hyperbolic space—well-suited for embedding tree-like data—as illustrated in Figure 1, to more effectively capture the underlying hierarchical relationships in image-text data. By introducing a Mixture-of-Experts model based on vision-text alignment, it more effectively manages the mapping relationships between vision, text, and actions, thereby enhancing its corresponding generalization capability.
\begin{figure}[t]\centering
	\includegraphics[width=\linewidth]{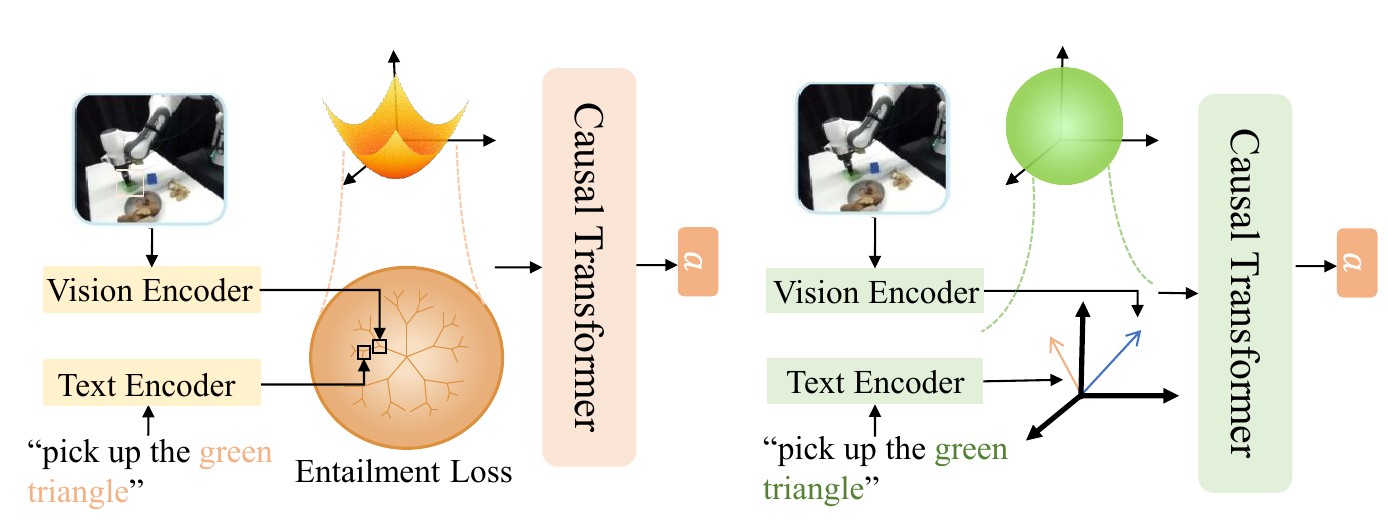}
    \caption{Comparison of hyperbolic representation (Left) and Euclidean contrastive Loss (Right) for VLA Alignment.}
    \label{pic:pro_attention}
\end{figure}

\begin{figure*}[t]\centering
	\includegraphics[width=\linewidth]{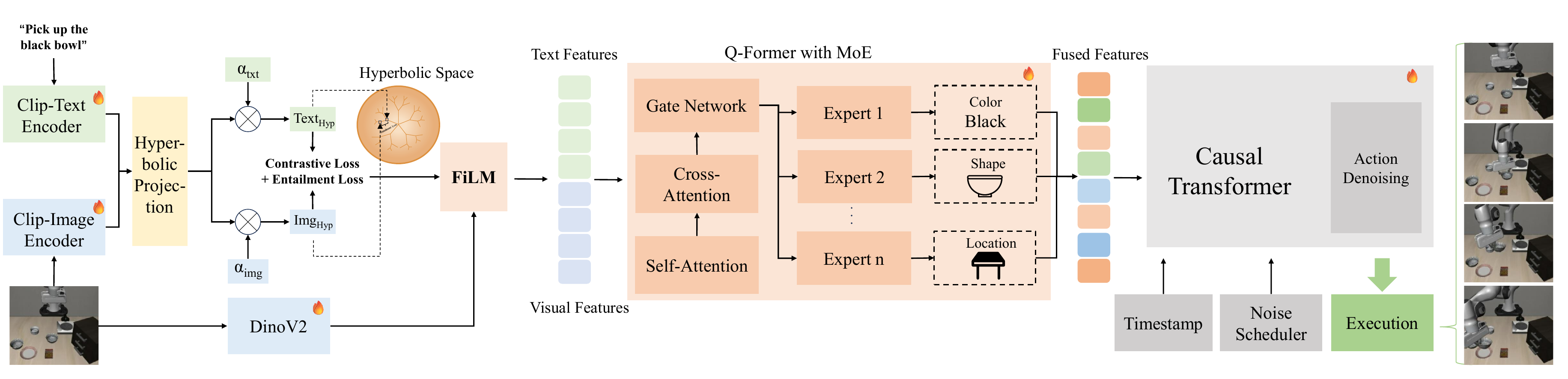}
	\caption{Overview of our HMVLA framework. A hierarchical VLA transformer with hyperbolic projection and a sparsely-gated Mixture-of-Experts (MoE) for enhanced semantic alignment.}
    \label{fig:struction}
\end{figure*}

In this paper, we propose HMVLA, a hyperbolic-based multimodal Vision-Language-Action (VLA) framework. Unlike existing approaches that primarily transfer prior knowledge from generative models or rely on reasoning strategies like chain-of-thought for image inference, our method fine-tunes the VLM while embedding multimodal features in hyperbolic space~\cite{chen1974hyperbolic}. As illustrated in Fig.~\ref{pic:pro_attention}, hyperbolic space, with its property of \textit{exponential expansion}, provides a natural representation for hierarchical or tree-like structures, enabling VLA models to capture underlying semantic hierarchies in vision–language data with minimal distortion~\cite{desai2023hyperbolic}. To further enhance multimodal fusion, we integrate a sparsely-gated Mixture-of-Experts (MoE) mechanism. By dynamically routing vision and language representations within hyperbolic space to specialized experts, the MoE facilitates adaptive alignment between modalities and actions, improving generalization. 
Our contributions are as follows:
\begin{itemize}[itemsep=0pt, topsep=0pt, leftmargin=1.5em]
    \item We introduce hyperbolic geometry into the VLA domain, leveraging its structural properties to better preserve hierarchical relationships during multimodal fusion.
    \item We incorporate the MoE module that adaptively routes information across experts, enhancing alignment between visual and linguistic modalities for action prediction.
    \item  Experiments show the effectiveness of our framework. We reconstruct the four datasets of the LIBERO~\cite{liu2023libero} to systematically verify the generalization capability of relevant models.
\end{itemize}

\section{Method}
\label{sec:pagestyle}
% In this section, we describe HMVLA in detail. We first introduce the hyperbolic modeling component \ref{HS}, which provides a structured representation space for multimodal inputs. We then present the mixture-of-experts mechanism \ref{MoE}, which enhances scalability and adaptivity by dynamically routing information across expert modules.
% In this section, we present our 
% We model the $n$-dimensional hyperbolic space as the upper sheet of a two-sheeted hyperboloidin $\mathbb{R}^{n+1}$.
\subsection{Hyperbolic Semantic Alignment}
\label{HS}
% The overall pipeline involves first independently encoding images and text using their respective encoders to obtain embedding vectors of dimension n. These embeddings are then transformed from Euclidean space into the Lorentz hyperbolic space. Finally, within this space, training objectives are introduced to incorporate both semantic information and structural priors from the images.
Our HMVLA framework is based on the Lorentz model in hyperbolic geometry, as shown in Figure~\ref{fig:struction}. Specifically, the $n$-dimensional hyperbolic space is represented as the upper sheet of a two-sheeted hyperboloid embedded in $\mathbb{R}^{n+1}$. Following the terminology from special relativity~\cite{einstein1905elektrodynamik}, we treat the hyperboloid’s axis of symmetry as the time dimension and the remaining axes as space dimensions. Thus, any vector $x \in \mathbb{R}^{n+1}$ can be expressed as $x = [x_{{space}}, x_{{time}}]$, where $x_{{space}} \in \mathbb{R}^n$ and $x_{{time}} \in \mathbb{R}$. The Lorentzian inner product is defined as:

% Let $⟨\cdot,\cdot⟩_{\mathcal{L}}$ denote the Lorentzian inner product that is induced by the Riemannian metric of the Lorentz model. For two vectors $x, y \in \mathbb{R}^{n+1}$, it is conputed as fellows:
% The Lorentz model with constant negative curvature $-c$ is defined by the following set of vectors:
% The Lorentzian inner product is：
 % The Lorentz model with curvature $-c$ is
\begin{equation}
    ⟨x,y⟩_{\mathcal{L}} = ⟨x_{space}, y_{space}⟩ - x_{time} y_{time}
\end{equation}
The induced Lorentzian norm is $||x||_{\mathcal{L}}=\sqrt{|⟨x, x⟩_{\mathcal{L}}|}$. 
The Lorentz model with curvature $-c$ is:
\begin{equation}
    \mathcal{L}^{n} = \{ x \in \mathbb{R}^{n+1}: ⟨x, x⟩_{\mathcal{L}}= -\frac{1}{c}\}, c > 0
\end{equation}
All vectors in $\mathcal{L}^n$ satisfy the following constraint:
\begin{equation}
    x_{time} = \sqrt{1/c+||x_{space}||^2}
\end{equation}
\textbf{Tangent Space} at $z \in \mathcal{L}^{n}$ consists of all vectors orthogonal to $z$ under the Lorentzian inner product, forming a Euclidean space:
\begin{equation}
    \mathcal{T}_{z}\mathcal{L}^{n} = \{v\in \mathbb{R}^{n+1}:⟨z, v⟩_\mathcal{L} = 0\}
\end{equation}
\textbf{Exponential and logarithmic map} provides a way to map vectors from the tangent space onto the manifold. For a point $z \in \mathcal{L}^n$, it is defined as:
\begin{equation}
    expm_{z}(v) = cosh(\sqrt{c}||v||_{\mathcal{L}})z + \frac{sinh(\sqrt{c}||v||_{\mathcal{L}})}{\sqrt{c}||v||_{\mathcal{L}}}
\end{equation}
The inverse operation is the logarithmic map (($logm_{z}$ : $\mathcal{L}^{n} \xrightarrow{} \mathcal{T}_z\mathcal{L}^n$), which maps a point $x$ on the hyperboloid back into the tangent space:
\begin{equation}
    logm_z(x)=\frac{cosh^{-1}(-c⟨z, x⟩_{\mathcal{L}})}{\sqrt{(c⟨z, x⟩_{\mathcal{L}})^{2}}} proj_z(x)
\end{equation}
We only consider these maps where $z$ is the origin of the hyperboloid ($\mathbf{O} = [0, p1/c]$).

Based on the relevant background of hyperbolic space discussed above, we now incorporate it into the HMVLA model. First, we obtain embedding vectors through the image and text encoders, and then apply linear projection to get $v_{enc}\in \mathbb{R}^{n}$ and $l_{enc} \in \mathbb{R}^{n}$. We then apply a transformation such that the vector $v_{enc}$ and $l_{enc}$ lie on the Lorentz hyperboloid $\mathcal{L}^{n}$ in $\mathbb{R}^{n+1}$. Let the vector $v=[v_{enc},0] \in \mathbb{R}^{n+1}$ and $l=[l_{enc}, 0] \in \mathbb{R}^{n+1}$.

% The tangent space at $z \in \mathcal{L}^{n}$ consists of all vectors orthogonal to $z$ under the Lorentzian inner product, forming a Euclidean space:
% \begin{equation}
%     \mathcal{T}_{z}\mathcal{L}^{n} = \{v\in \mathbb{R}^{n+1}:⟨z, v⟩_\mathcal{L} = 0\}
% \end{equation}

 We find that $v$ and $l$ belong to tangent space at the hyperboloid origin $\mathbf{O}$ as Eq.4 is satisfied: $⟨\mathbf{O}, v⟩_{\mathcal{L}}=0$ and $⟨\mathbf{O}, l⟩_{\mathcal{L}}=0$. Therefore, we parameterize only the spatial components of the Lorentz model $(v_{enc} = v_{space}$ and $l_{enc} = l_{space})$.
%  The exponential map provides a way to map vectors from the tangent space onto the manifold. For a point $z$ on the hyperboloid, it is defined as $expm_z : \mathcal{T}_z\mathcal{L}^{n} \xrightarrow {} \mathcal{L}^{n}$, with the specific expression:
% \begin{equation}
%     expm_{z}(v) = cosh(\sqrt{c}||v||_{\mathcal{L}})z + \frac{sinh(\sqrt{c}||v||_{\mathcal{L}})}{\sqrt{c}||v||_{\mathcal{L}}}
% \end{equation}
We can simplify the exponential map in Eq.5 through the above parameterization approach, with the formula given as follows:
\begin{equation}
    y_{space} = cosh(\sqrt{c}||v||_{\mathcal{L}})0 + \frac{sinh(\sqrt{c}||v||_{\mathcal{L}})}{\sqrt{c}||v||_{\mathcal{L}}} v_{space}
\end{equation}
\begin{equation}
    x_{space} = cosh(\sqrt{c}||l||_{\mathcal{L}})0 + \frac{sinh(\sqrt{c}||l||_{\mathcal{L}})}{\sqrt{c}||l||_{\mathcal{L}}} l_{space}
\end{equation}
The Lorentzian norm of vector $v$ and $l$ simplifies to the Euclidean norm of its spatial components, and can therefore be reduced to:
\begin{equation}
    y_{space} =\frac{sinh(\sqrt{c}||v_{space}||)}{\sqrt{c}||v_{space}||} v_{space}
\end{equation}
\begin{equation}
    x_{space} =  \frac{sinh(\sqrt{c}||l_{space}||)}{\sqrt{c}||l_{space}||} l_{space}
\end{equation}
For the traditional CLIP contrastive loss, we incorporate an entailment loss to enhance the structural relationships between text and images. $y_{time}$ and $x_{time}$ are calculated using Eq.3. To this end, we further impose an entailment cone constraint in the hyperbolic space to model such hierarchical dependency. This cone is defined by the half-aperture:
\begin{equation}
    aper(x) = sin^{-1}(\frac{2K}{\sqrt{c}||x_{space}||})
\end{equation}
where a constant $K$ represents boundary conditions near the origin. We now aim to identify and penalize when the paired image embedding $y$ lies outside the entailment cone. We measure the exterio angle $ext(x,y) = \pi - \angle{Oxy}$:
% and define the exterior angle $ext(x,y) = \pi - \angle{Oxy}$:
\begin{equation}
    ext(x,y) = cos^{-1}(\frac{y_{time}+x_{time}c⟨x,y⟩_{\mathcal{L}}}{||x_{space}||\sqrt{(c⟨x,y⟩_{\mathcal{L}})^2}-1})
\end{equation}
If the exterior angle is smaller than the aperture, it indicates that the constraint relationship between $x$ and $y$ is satisfied; otherwise, the angular difference needs to be reduced through a loss function:
\begin{equation}
    \mathcal{L}_{ent}(x,y) = max(0, ext(x,y)-aper(x))
\end{equation}
Finally, the overall objective is:
\begin{equation}
\mathcal{L} = \mathcal{L}_{\text{cont}} + \lambda \mathcal{L}_{\text{ent}}
\end{equation}
where $\lambda$ = 0.1 is a balancing coefficient controlling the relative importance of the two losses. 

\subsection{Soft MoE for Multimodal Fusion}
\label{MoE}

To better capture fine-grained image-text semantics, we enhance the Q-Former by replacing its feed-forward layers with a soft Mixture-of-Experts (softMOE) module. The Q-Former first encodes visual and textual features through self- and cross-attention, producing fused query tokens $\{q_i\}_{i=1}^N$, which are then processed by the MoE.

The MoE consists of $M$ expert networks $\{E_m\}_{m=1}^M$ and a gating network $G(\cdot)$. For each token $q_i$, the gating network outputs logits $g_i \in \mathbb{R}^M$, normalized via softmax as:
\begin{equation}
w_i^m = \frac{\exp(g_i^m)}{\sum_{j=1}^M \exp(g_i^j)}
\end{equation}
where the token is updated by:
\begin{equation}
\tilde{q}_i = \sum_{m=1}^M w_i^m E_m(q_i)
\end{equation}
This soft routing lets each token access all experts while focusing on the most relevant ones.

We insert the softMOE into each Transformer block by replacing the feed-forward layer: attention outputs are normalized, processed by MoE, and merged via residual connections. To encourage balanced expert usage, we add a load-balancing loss:
\begin{equation}
\mathcal{L}_{\text{balance}}
= M \sum_{m=1}^M 
\left(\frac{1}{N}\sum_{i=1}^N w_i^m\right)
\left(\frac{n_m}{N}\right)
\end{equation}
where $n_m$ is the number of tokens routed to expert $m$. The final training objective is:
\begin{equation}
\mathcal{L}_{\text{MoE}} = \mathcal{L}_{\text{task}} + \beta \mathcal{L}_{\text{balance}}
\end{equation}
where $\beta$ controls the regularization strength. This design improves semantic alignment while maintaining efficiency.

\section{EXPERIMENTS}
\label{sec:typestyle}

\textbf{Datasets.}  We used the LIBERO~\cite{liu2023libero} benchmark for evaluation. Specifically, we utilized four datasets: Spatial, Object, Goal, and LONG. To validate the generalization capability of our model, we reconstructed a new dataset (Gen) for training and validation to examine its generalizability.

\textbf{Implementation Details.} We adopt Dita~\cite{hou2025dita} as the backbone network. Training is conducted for 80k steps using the Adam optimizer with a learning rate of $1\times10^{-4}$. The hyperbolic curvature is set to $0.1$, and the Mixture-of-Experts (MoE) module includes 6 experts. The model is trained with a batch size of 64 and input image resolutions of $256 \times 256$ pixels. For trajectory modeling, we set the trajectory length (traj\_length) to 11 and the trajectory dimension (trajectory\_dim) to 7. At each step, the model predicts 10 future actions. All experiments are performed on an NVIDIA H200 GPU.

\begin{figure}[!t] % 使用 [t] 保持和原来 table 相同位置
\centering
\includegraphics[width=1\linewidth]{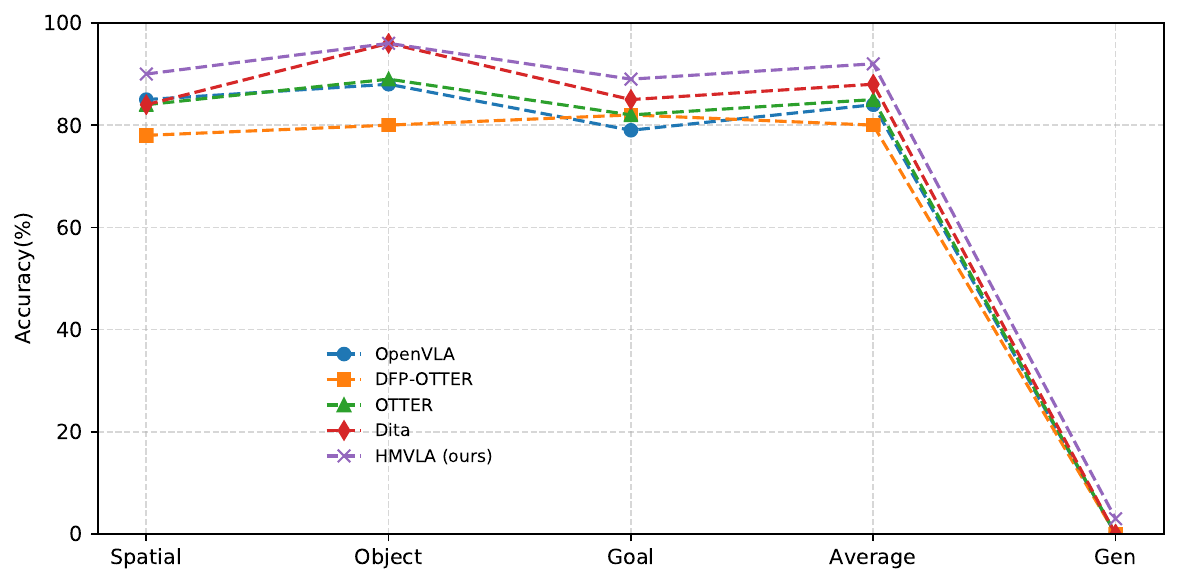} % 替换成你的 PDF 文件名
\caption{Comparison of HMVLA and Baseline Models on Spatial, Object, Goal, and Our Constructed Datasets.}
\vspace{-5mm}
\label{fig:methodresult}
\end{figure}

\textbf{Comparison with Advanced Methods.} As shown in Table~\ref{tab:metrics_comparison} and Fig.~\ref{fig:methodresult}, we conducted comparative experiments between our proposed method and state-of-the-art approaches.

\begin{table}[!t]
\centering
\renewcommand\arraystretch{1.3}
\setlength{\tabcolsep}{0.09cm}
\caption{Comparison of task-level accuracy of existing methods and our HMVLA model on LIBERO benchmark.}
\label{tab:metrics_comparison}
\scalebox{1}{
\begin{tabular}{ccccccccccccc}
\cline{1-13}
Method 
                         & \multicolumn{2}{c}{Spatial} & \multicolumn{2}{c}{Object}    & \multicolumn{2}{c}{Goal}      & \multicolumn{2}{c}{LONG}     & \multicolumn{2}{c}{Average}      &   
 &    \cline{1-1} \hline
DP~\cite{chi2023diffusion}          & \multicolumn{2}{c}{78\%}                        & \multicolumn{2}{c}{92\%}    &        \multicolumn{2}{c}{68\%}     &  \multicolumn{2}{c}{50\%}    &  \multicolumn{2}{c}{72\%} \\
Octo~\cite{team2024octo}          & \multicolumn{2}{c}{79\%}                          & \multicolumn{2}{c}{86\%}    &        \multicolumn{2}{c}{85\%}           &  \multicolumn{2}{c}{50\%}    &  \multicolumn{2}{c}{75\%}                   \\
Tra-MoE~\cite{yang2025tra}         & \multicolumn{2}{c}{69\%}                          & \multicolumn{2}{c}{77\%}    &        \multicolumn{2}{c}{88\%}     &    \multicolumn{2}{c}{31\%}     &    \multicolumn{2}{c}{66\%}                  \\
CoT-VLA~\cite{zhao2025cot}            & \multicolumn{2}{c}{87\%}                          & \multicolumn{2}{c}{91\%}    &        \multicolumn{2}{c}{87\%}     &    \multicolumn{2}{c}{69\%}               &    \multicolumn{2}{c}{81\%}            \\
Dita~\cite{hou2025dita}             & \multicolumn{2}{c}{84\%}                          & \multicolumn{2}{c}{96\%}    &        \multicolumn{2}{c}{85\%}     &    \multicolumn{2}{c}{63\%}      &        \multicolumn{2}{c}{82\%}                    \\
HMVLA \textbf{(ours)}     & \multicolumn{2}{c}{90\%}                          & \multicolumn{2}{c}{96\%}    &        \multicolumn{2}{c}{89\%}     &    \multicolumn{2}{c}{69\%}             &    \multicolumn{2}{c}{86\%}          \\ \cline{1-13}
\end{tabular}
}
\end{table}

\begin{figure}[!t] % 使用 [t] 保持和原来 table 相同位置
\centering
\includegraphics[width=1\linewidth]{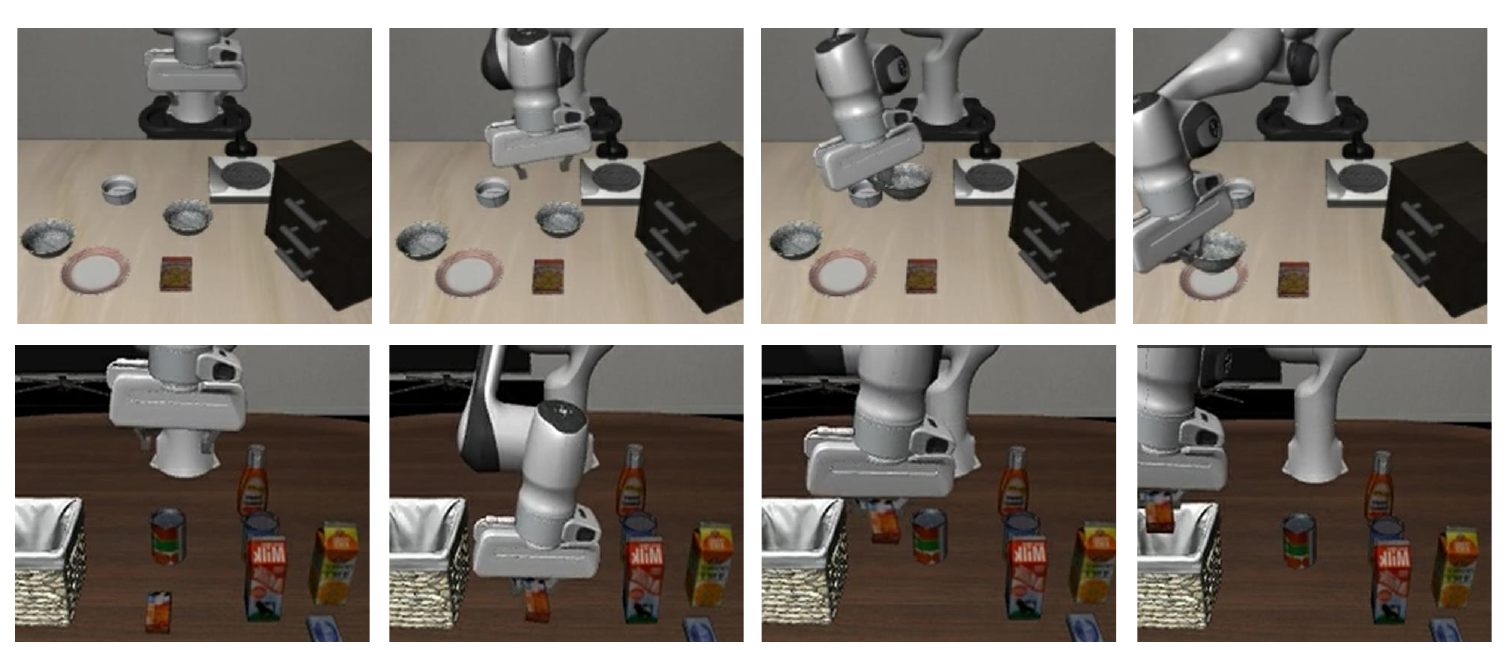} % 替换成你的 PDF 文件名
\caption{The HMVLA model's grasping process on LIBERO benchmark.}
\label{fig:example}
\end{figure}

As shown in Table~\ref{tab:metrics_comparison} and Fig.~\ref{fig:methodresult}, we conducted comparative experiments between the proposed method and state-of-the-art approaches. Table~\ref{tab:metrics_comparison} includes comparisons with DP~\cite{chi2023diffusion}, Octo~\cite{team2024octo}, Tra-MoE~\cite{yang2025tra}, CoT-VLA~\cite{zhao2025cot}, and Dita~\cite{hou2025dita} algorithms, where our framework demonstrates superior performance. Figure~\ref{fig:methodresult} further validates the generalization capability of our model, with comparative models including OpenVLA~\cite{kim2024openvla}, DFP-OTTER~\cite{huang2025otter}, OTTER~\cite{huang2025otter}, and Dita~\cite{hou2025dita}. And Fig.~\ref{fig:example} shows the grasping process of our HMVLA model. 

Our method not only achieves the best performance on the benchmark dataset but also exhibits breakthrough improvements on the generalization validation set compared to other models. Compared to other algorithms, our model leverages the inherent tree-like architecture of hyperbolic space to naturally align text and image embeddings, while the softMoE mechanism is responsible for modeling specific semantic components within instructions, thereby enhancing the decomposition and generalization capabilities of cross-modal semantics.

\begin{figure}[!t]
\centering
\includegraphics[width=1\linewidth]{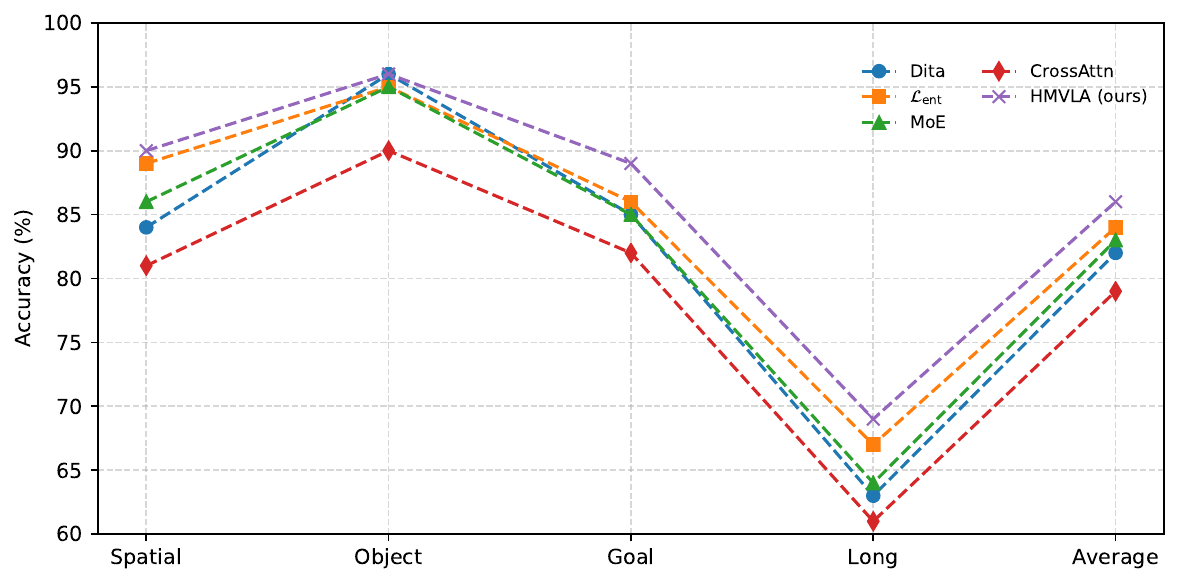}  % 替换为你的 PDF 文件名
\caption{Ablation Experiments Evaluating Hyperbolic, MoE, and Cross-Attention.}
\vspace{-5mm} % 调整图题与下方正文间距（可正可负）
\label{fig:ablation}
\end{figure}

\textbf{Ablation Study.} As shown in Fig.~\ref{fig:ablation}, $\mathcal{L}_{ent}$ constrains hyperbolic space during modality fusion, while MoE denotes that different experts can be scheduled under the routing mechanism. We also tested replacing FiLM with Cross-Attention. We observe that introducing hyperbolic space into Dita~\cite{hou2025dita} significantly improves task success, especially for complex instructions where the original Dita model struggled. This demonstrates that hyperbolic space effectively captures hierarchical structure, enhancing multimodal fusion. In addition, the routing ability of MoE helps avoid overfitting to a single fused representation, while FiLM provides more stable and balanced modality conditioning than Cross-Attention. Together, these results confirm that each component contributes to more robust semantic-to-action alignment.
\section{Conclusion}
\label{sec:pagestyle}
In this paper, we propose HMVLA, a novel Vision-Language-Action (VLA) model that leverages hyperbolic space representation and a sparsely-gated Mixture-of-Experts (MoE) mechanism to enhance semantic alignment between vision and language for robotic control. By embedding multimodal features into a hyperbolic space, our model effectively captures the inherent hierarchical relationships within image-text data. Furthermore, the incorporation of MoE improves the decomposition and modeling of fine-grained semantic elements, leading to superior multimodal fusion. Experimental results on the LIBERO benchmark and our dataset demonstrate that HMVLA not only outperforms existing state-of-the-art methods in task accuracy but also exhibits strong generalization ability to novel instructions and previously unseen objects. 

% In future work, we aim to deploy HMVLA on real-world robotic platforms and further enhance its generalization ability to unseen objects, thereby improving the model’s reliability and adaptability in practical applications.

%\vfill\pagebreak

\section{ACKNOWLEDGEMENT}
This work was supported by the National Natural Science Foundation of China (Grant No. 62277011),  National Key Research and Development Program of China (Grant No. GG-2024-01-02), Project of Chongqing MEITC (Grant No. YJX-2025001001009) and Open Research Fund from Guangdong Laboratory of Artificial Intelligence and Digital Economy (SZ) (Grant No. GML-KF-24-18).

%\section{ACKNOWLEDGEMENT}
%\label{sec:refs}
%This work was supported by the National Key Research and Development Program of China (Grant No. GG-2024-01-02), National Natural Science Foundation of China (Grant No. 62277011) and Project of Chongqing MEITC(Grant No. YJX-2025001001009).

%\section{REFERENCES}
%\label{sec:refs}

% List and number all bibliographical references at the end of the
% paper. The references can be numbered in alphabetic order or in
% order of appearance in the document. When referring to them in
% the text, type the corresponding reference number in square
% brackets as shown at the end of this sentence \cite{C2}. An
% additional final page (the fifth page, in most cases) is
% allowed, but must contain only references to the prior
% literature.

% Please follow the IEEE Citation Guidelines, \url{https://ieee-dataport.org/sites/default/files/analysis/27/IEEE\%20Citation\%20Guidelines.pdf} for formatting of references.

%\begin{refcontext}[sorting=none]
%\printbibliography[heading=none]  % 禁止自动标题
%\end{refcontext}

\bibliographystyle{IEEEbib}\small
\bibliography{refs}

\end{document}